\documentclass[conference]{IEEEtran}
\IEEEoverridecommandlockouts

\usepackage{cite}
\usepackage[hidelinks]{hyperref}
\usepackage{amsmath,amssymb,amsfonts}
\usepackage{graphicx}
\usepackage{textcomp}
\usepackage{xcolor}
\usepackage{booktabs}
\usepackage{array}
\usepackage{pifont}
\usepackage{makecell}
\usepackage{multirow}
\usepackage{subcaption}
\usepackage{tabularx}
\usepackage{tikz}
\usepackage{algorithm}
\usepackage{algorithmic}
\usepackage{xspace}
\usepackage{glossaries}
\usepackage[capitalize,nameinlink,noabbrev]{cleveref}
\usetikzlibrary{shapes.geometric, arrows.meta, positioning}

\newcommand{\cmark}{\ding{51}}  % check
\newcommand{\xmark}{\ding{55}}  % cross
\newcommand{\simToReal}{\mbox{sim-to-real}\xspace}

\newcommand{\mAPfull}{\text{mAP}_{50\text{-}95}\xspace}

\newcommand{\refDataset}{\mathcal{D}_{\text{ref}}}
\newcommand{\trainDataset}{\mathcal{D}_{\text{train}}}
\newcommand{\testDataset}{\mathcal{D}_{\text{test}}}

\newacronym{HMLV}{HMLV}{High-Mix, Low-Volume}
\newacronym{DR}{DR}{Domain Randomization}
\newacronym{DA}{DA}{Domain Adaptation}
\newacronym{S2R}{sim-to-real}{Simulation-to-Reality}
\newacronym{SAL}{SAL}{Synthetic Active Learning}
\newacronym{S-GDR}{S-GDR}{Semantically-Guided Domain Randomization}
\newacronym{VLM}{VLM}{Vision-Language Model}
\newacronym{SDG}{SDG}{Synthetic Data Generation}
\newacronym{CAD}{CAD}{Computer-Aided Design}
\newacronym{SDXL}{SDXL}{Stable Diffusion XL}
\newacronym{LDM}{LDM}{Latent Diffusion Model}
\newacronym{IoU}{IoU}{Intersection-over-Union}
\newacronym{mAP}{mAP}{mean Average Precision}
\newacronym{AP}{AP}{Average Precision}
\newacronym{GDR}{GDR}{Guided Domain Randomization}

\begin{document}

\title{Semantically-Guided Domain Randomization for Industrial Object Detection in Low-Image-Budget Regimes\\
\thanks{%This project is conducted at the ARENA2036 research campus. It is
funded by the Ministry of Economic Affairs, Labour and Tourism Baden Württemberg
as part of the Industrial Metaverse initiative.\\
\textsuperscript{*}Corresponding author.
}}

\author{%
\IEEEauthorblockN{1\textsuperscript{st} Jose Moises Araya-Martinez\textsuperscript{*}}
\IEEEauthorblockA{\textit{Electrical Engineering and Computer Science}\\
\textit{TU Berlin}\\
Berlin, Germany\\
araya.martinez@campus.tu-berlin.de}
\and
\IEEEauthorblockN{2\textsuperscript{nd} Gautham Mohan}
\IEEEauthorblockA{\textit{Electrical Engineering}\\
\textit{University of Stuttgart}\\
Stuttgart, Germany\\
st184914@stud.uni-stuttgart.com}
\and
\IEEEauthorblockN{3\textsuperscript{rd} Jens Lambrecht}
\IEEEauthorblockA{\textit{Institute for Cognitive Robotics}\\
\textit{TU Braunschweig}\\
Braunschweig, Germany\\
jens.lambrecht@tu-braunschweig.de}
}

\maketitle

% ==========================================================================
\begin{abstract}
Retraining visual perception pipelines in \acrfull{HMLV} automotive manufacturing must be carried out under tight annotation, energy, and time budgets, yet most \acrfull{SDG} strategies still operate in the thousands of images. This work evaluates \acrfull{S-GDR}, an annotation-free adaptation pipeline that couples \acrfull{VLM}-based semantic captioning of a small unannotated real reference set with diffusion-based background synthesis (\acrfull{SDXL} conditioned by ControlNet and IP-Adapter) and mask-based object composition. On an automotive multi-object detection benchmark and with a fixed budget of 200 synthetic training images, S-GDR reaches $ \mAPfull = 0.739$ on a real held-out test set, outperforming a domain-randomized render baseline ($\mAPfull = 0.697$) as well as brightness filtering, perceptual hashing, CycleGAN style transfer, and unguided diffusion variants sharing the same 200-image budget. These initial observations position S-GDR as a promising annotation-free alternative for extreme data-scarcity regimes.
\end{abstract}


% ==========================================================================
\section{Introduction}
\label{sec:intro}
% ==========================================================================
Visual perception underpins factory automation and smart manufacturing, yet
deep learning models for object detection, segmentation, and pose estimation
still rely on large, manually annotated
datasets~\cite{demlehner2020shall}. This burden is particularly acute in
\acrfull{HMLV} settings~\cite{hansen2021artificial}, where high product variety
demands frequent retraining and perception pipelines must scale automatically
to track real-world variability~\cite{prakash2019structured}.
In automotive body-in-white assembly, for example, a single perception task
can involve many visually similar sheet-metal parts, industrial containers, and
subassemblies whose appearance changes across model years, plants, and
lighting conditions.

This paper focuses on the practical regime in which
(i)~the task is multi-object detection of automotive parts,
(ii)~no manual annotations are available,
(iii)~a small unannotated set of real deployment images can be collected
(order of tens of images), and
(iv)~the number of synthetic training images is capped at a few
hundred (in our experiments, $|\trainDataset|=200$) for
financial, energy, and iteration-time reasons~\cite{Strubell2019}.
Under this budget, the research question we address is:
\emph{can grounding synthetic image generation on semantic descriptions
of the real deployment context produce higher-value training data than
randomization or feature-based selection~?}

\begin{figure*}[t]
    \centering
    \includegraphics[width=0.95\textwidth]{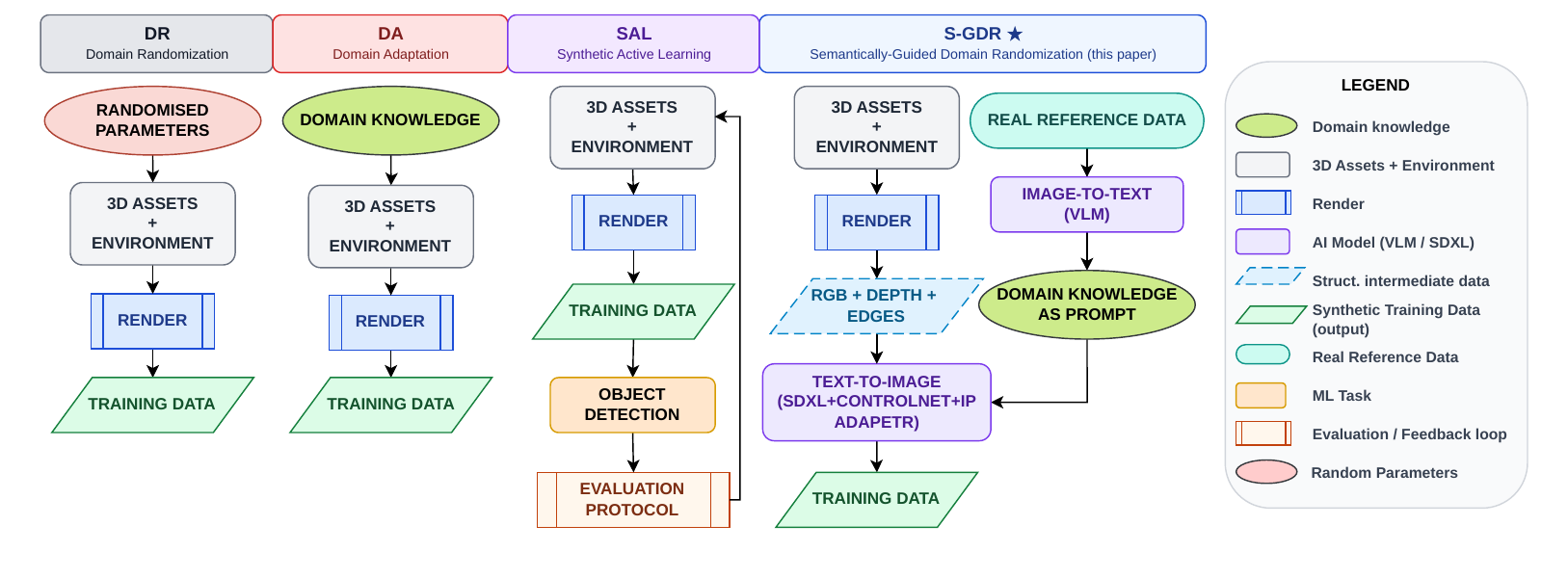}
    \caption{Comparison of \acrshort{SDG} paradigms.
    \acrshort{DR} randomizes simulation parameters in an open loop;
    \acrshort{DA} incorporates domain knowledge via manual tuning;
    \acrshort{SAL} closes the loop with a fixed, non-semantic evaluator on
    synthetic data; the proposed \acrshort{S-GDR} replaces the fixed evaluator
    with \acrshort{VLM}-driven semantic feedback over real reference data
    $\refDataset$ and conditions a generative model on the resulting caption
    to produce context-aware training images.}
    \label{fig:comparison_of_SDG_methods}
\end{figure*}

\acrfull{SDG} has emerged as a key strategy to reduce the annotation
burden. As illustrated in \cref{fig:comparison_of_SDG_methods}, existing
approaches span four broad paradigms.
\acrfull{DR}~\cite{tobin2017domain,Tremblay2018,Zhu2025icra} varies
simulation parameters in an open loop, often producing task-irrelevant
samples~\cite{araya-martinez2025xai}.
\acrfull{DA} incorporates domain knowledge via manual tuning and structured
randomization~\cite{prakash2019structured,Mayershofer2021,arayamartinez2026synthrenderirisopensourceframework},
but does not scale automatically to novel environments without human
intervention.
\acrfull{SAL}~\cite{ZHU202668} introduces feedback-driven generation, yet
evaluates distributional gaps using fixed criteria on synthetic data, leaving
the \simToReal gap of generated samples unverified against a real target
distribution.

Prior work~\cite{araya-martinez2025genai} has shown that when sufficient
rendering variability is available (a few thousand images), simple
feature-based methods aligning synthetic and real distributions can outperform
generative augmentation in both accuracy and compute cost. In parallel,
GenAI-based 3D asset reconstruction has been proposed as a viable
\acrshort{DA} alternative to manual geometric part
modeling~\cite{arayamartinez2026synthrenderirisopensourceframework}. The
question we investigate here is complementary: whether generative augmentation
provides higher-value contextualized data specifically in the
\emph{extreme data-scarcity regime} that HMLV retraining forces on
manufacturers.

Concretely, we generalise the pipeline of~\cite{araya-martinez2025genai}
as \acrfull{S-GDR}: a family of methods that couples \acrshort{DR} with
\acrshort{VLM}-based semantic scene evaluation of the real reference set
and with diffusion-based image synthesis conditioned by that semantic
description. The working hypothesis is that, under a fixed 200-image budget,
combining a high-variability randomized subset with a semantically
contextualized counterpart produces stronger multi-target object detection
performance than randomization alone or than feature-based selection
methods, while remaining fully annotation-free.

The main contributions of this work-in-progress paper are:
\begin{itemize}
    \item We instantiate and evaluate \acrshort{S-GDR} using
    Qwen2\nobreakdash-VL~\cite{wang2024qwen2vl} as the semantic
    captioner and \acrshort{SDXL}~\cite{podell2023sdxl} with
    ControlNet~\cite{zhang2023adding} and
    IP-Adapter~\cite{ye2023ipadapter}, followed by mask-based
    object composition, all under a fixed budget of 200 synthetic
    training images.
    \item On the public automotive multi-object detection benchmark
    of~\cite{martinez2024scap}, S-GDR obtains
    $\mAPfull = 0.739$ on the real held-out test set, versus a
    domain-randomized render baseline of $0.697$ and four
    additional feature-based and generative variants that share the
    same 200-image budget.
    \item We position \acrshort{S-GDR} against the \acrshort{SDG}
    landscape (\cref{tab:sdg_comparison}), and we discuss the
    limitations of the current single-run, single-detector,
    single-benchmark evaluation, including the effect of mask-based
    composition and possible blending
    alternatives~\cite{perez2003poisson,tsai2017deepharmonization,ghiasi2021copypaste},
    the qualitative computational cost of the pipeline, and the
    open ablation of the individual conditioning components.
\end{itemize}

% ==========================================================================
\section{Related Work}
\label{sec:related_works}
% ==========================================================================
This section only complements \cref{sec:intro}; the four \acrshort{SDG}
paradigms in \cref{fig:comparison_of_SDG_methods} are not restated.

\subsection{Data Scarcity in Industrial Vision}
Foundation models for segmentation~\cite{Kirillov_2023_ICCV} and 6D pose
estimation~\cite{wen2024foundationpose} reduce data dependency but still
require a prior detector for downstream tasks.
Zero-shot detectors such as Grounding DINO~\cite{liu2023grounding} remain
inferior to supervised counterparts on domain-specific industrial objects,
motivating the practical need for automated data generation to enable
retraining in \acrshort{HMLV}
settings~\cite{hansen2021artificial,prakash2019structured}.

\subsection{Domain Randomization and Domain Adaptation for SDG}
\acrshort{DR}~\cite{tobin2017domain,Tremblay2018,Zhu2025icra} generates
annotated images by randomizing simulation parameters over wide ranges;
without feedback from the target domain, it may produce task-irrelevant
samples and offers no guarantee of distributional
coverage~\cite{araya-martinez2025xai}. Structured
\acrshort{DA}~\cite{prakash2019structured,Mayershofer2021} and
photorealistic asset
reconstruction~\cite{arayamartinez2026synthrenderirisopensourceframework}
constrain synthesis using domain knowledge but require iterative human
tuning and do not generalize automatically to new environments.
\acrshort{SAL}~\cite{ZHU202668} closes the loop by triggering data
generation in response to model evaluation, but its evaluators are fixed,
non-semantic, and applied to synthetic rather than real data.

\subsection{Generative and Semantic Adaptation}
Latent diffusion models such as Stable
Diffusion~\cite{Rombach_2022_CVPR} and
\acrshort{SDXL}~\cite{podell2023sdxl} enable photorealistic image
synthesis conditioned on text prompts, depth maps, or reference images.
Spatial conditioning is typically added via
ControlNet~\cite{zhang2023adding}, and appearance conditioning via
IP-Adapter~\cite{ye2023ipadapter}. Diffusion-based data augmentation for
downstream vision tasks has been explored, e.g., by
DA-Fusion~\cite{trabucco2024effective}, but not specifically for
industrial multi-object detection in the 200-image regime.
Prior work~\cite{araya-martinez2025genai} has shown that with a few
thousand rendered images simple feature-based methods can match or
outperform generative augmentation. This paper investigates the
complementary regime.
\acrfull{VLM}s such as LLaVA~\cite{liu2023llava} and
Qwen2-VL~\cite{wang2024qwen2vl} provide open-vocabulary natural-language
descriptions of visual scenes; their use as \emph{semantic feedback
signals} within an \acrshort{SDG} pipeline to condition generative models
on real deployment context is, to our knowledge, unexplored for
industrial object detection.

\subsection{Object Compositing}
S-GDR composites rendered target objects onto generated backgrounds using
their segmentation masks. Simple copy-paste has been shown to be a strong
augmentation for instance segmentation~\cite{ghiasi2021copypaste}, but is
known to introduce photometric discontinuities at object boundaries when
foreground and background statistics differ.
Poisson image editing~\cite{perez2003poisson} and learned
harmonization~\cite{tsai2017deepharmonization} attenuate these
inconsistencies at the cost of additional processing.
\cref{sec:discussion} revisits this trade-off in the context of the
present pipeline.

\subsection{Positioning S-GDR in the SDG Landscape}
\cref{tab:sdg_comparison} locates \acrshort{S-GDR} against prior work
along dimensions critical to data-scarce applications. The key
differentiator is the training set scale: whereas most \acrshort{SDG}
methods operate in the thousands of images to reach higher terminal
accuracies, S-GDR, derived from~\cite{araya-martinez2025genai}, targets
the 200-image regime while introducing semantic, real-data-grounded
feedback.

\begin{table}[t]
\centering
\caption{Representative \acrshort{SDG} approaches for object detection.
\cmark~=~present; \xmark~=~absent. Adaptation features:
\textit{low}~=~color/lighting/texture; \textit{struct.}~=~geometry/layout;
\textit{sem.}~=~natural-language semantics.}
\label{tab:sdg_comparison}
\resizebox{\columnwidth}{!}{%
\begin{tabular}{lllccc}
\toprule
\textbf{Work} &
\textbf{Strategy} &
\textbf{DA\ Features} &
\textbf{Real Context} &
\textbf{Sem.\ Eval.} &
\textbf{Train Set} \\
\midrule
Tobin et al.~\cite{tobin2017domain}                  & \acrshort{DR}  & None            & \xmark & \xmark & $\geq$5k \\
Tremblay et al.~\cite{Tremblay2018}                  & \acrshort{DR}  & Low             & \xmark & \xmark & $\geq$2.5k \\
Prakash et al.~\cite{prakash2019structured}          & \acrshort{DA}  & Struct.         & \xmark & \xmark & $\geq$1k \\
Mayershofer et al.~\cite{Mayershofer2021}            & \acrshort{DA}  & Low + struct.   & \xmark & \xmark & $\geq$2.5k \\
SynthRender~\cite{arayamartinez2026synthrenderirisopensourceframework}
                                                     & \acrshort{DA}  & Low + struct.   & \xmark & \xmark & $\geq$400 \\
Zhu et al.~\cite{ZHU202668}                          & \acrshort{SAL} & Low (synth.)    & \xmark & \xmark & $\geq$4k \\
Araya-Martinez et al.~\cite{araya-martinez2025genai} & \acrshort{GDR} & Low + sem.\ (part.) & \cmark & Partial & $\geq$400 \\
\midrule
\textbf{S-GDR (ours)} & \textbf{\acrshort{S-GDR}} & \textbf{Sem.\ (VLM)} & \cmark & \cmark & \textbf{200} \\
\bottomrule
\end{tabular}%
}
\end{table}

% ==========================================================================
\section{Methodology}
\label{sec:methodology}
% ==========================================================================

\subsection{Problem Setting, Detector, and Dataset}
\label{sub:data_model_and_train_regime}

\begin{figure}[t]
\centering
\begin{minipage}[b]{1\linewidth}
    \centering
    \subcaptionbox{}{\includegraphics[width=0.32\linewidth]{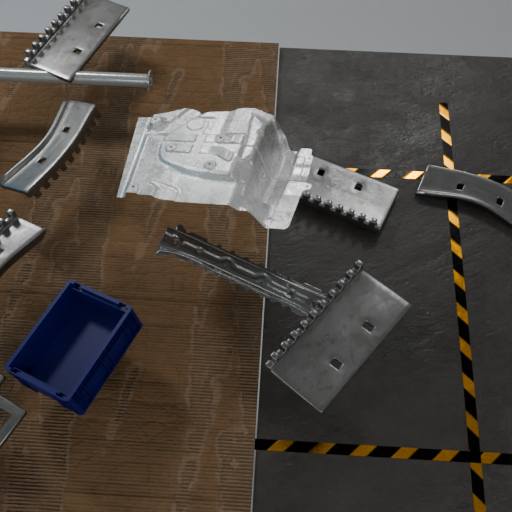}}\hfill
    \subcaptionbox{}{\includegraphics[width=0.32\linewidth]{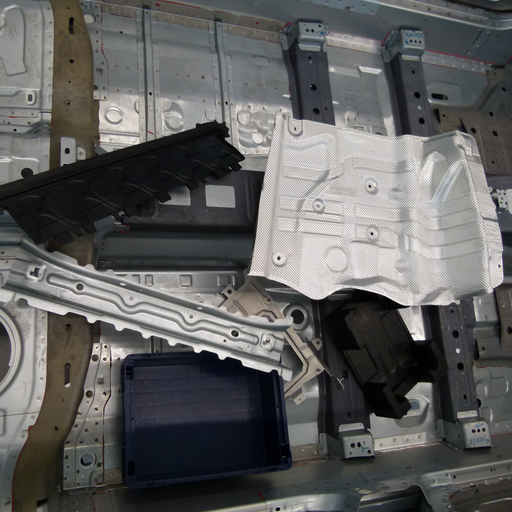}}\hfill
    \subcaptionbox{}{\includegraphics[width=0.32\linewidth]{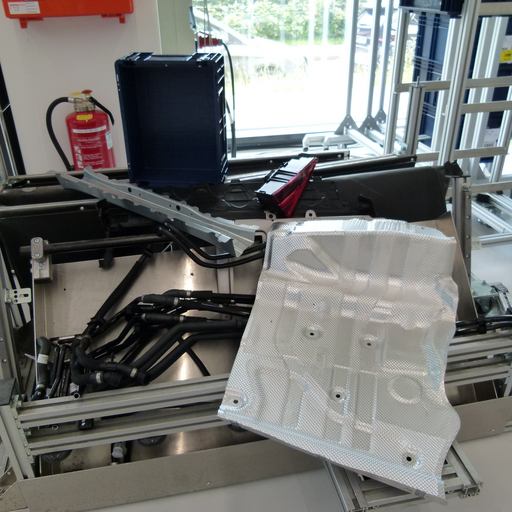}}
\end{minipage}
\caption{Representative samples from the three dataset partitions used in
this work: (a) domain-randomized rendered training image
($\trainDataset$); (b) real test image ($\testDataset$), used solely for
evaluation; (c) real reference image ($\refDataset$), used as semantic
context source for \acrshort{S-GDR} augmentation.}
\label{fig:datasets}
\end{figure}

We use the public automotive multi-object detection benchmark introduced
in~\cite{martinez2024scap}, which contains three disjoint image sets
illustrated in \cref{fig:datasets}:
\begin{itemize}
    \item a synthetic training set $\trainDataset$ of domain-randomized
    rendered images produced from \acrshort{CAD} models of the target
    parts;
    \item a real-world test set $\testDataset$ used \emph{exclusively}
    for performance evaluation; and
    \item a small real-world reference set $\refDataset$ used by
    \acrshort{S-GDR} as the semantic context source for generative
    augmentation.
\end{itemize}
The task is multi-class detection of automotive body-in-white
components (denoted in \cref{fig:200_img_map_results} as \emph{Part~0},
\emph{Part~0.1}, and \emph{Industrial Box}). Training uses only synthetic
images; images from $\testDataset$ and $\refDataset$ are excluded from
training, so all reported detection scores refer to $\testDataset$.

All experiments use YOLOv8~\cite{Jocher_Ultralytics_YOLO_2023} with
default hyperparameters and no task-specific tuning, in order to
isolate the effect of training-data composition from model
optimization. Under this fixed detector setting, differences between
methods can be attributed to the training data alone, but only within
the range of behaviours the default-hyperparameter YOLOv8 exposes; we
return to this limitation in \cref{sec:limitations}. All evaluations use
the standard COCO-style~\acrfull{mAP} averaged over
\acrshort{IoU} thresholds from $0.50$ to $0.95$ in steps of $0.05$,
denoted $\mAPfull$, and computed on $\testDataset$.

\subsection{The S-GDR Pipeline}
\label{sub:sgdr_pipeline}

\acrshort{S-GDR} generalizes the methodology
of~\cite{araya-martinez2025genai} as a family of annotation-free
adaptation methods that extract contextual information from
$\refDataset$ to automatically augment $\trainDataset$. The concrete
instantiation evaluated here is depicted in
\cref{fig:comparison_of_SDG_methods} and detailed in
\cref{alg:sgdr}. It has three stages.

\paragraph{Semantic captioning}
For each reference image $r\in\refDataset$, Qwen2\nobreakdash-VL~\cite{wang2024qwen2vl}
produces a natural-language description of the scene. In the
\emph{guided-prompt} configuration, the resulting captions are used as
positive prompts $p$ conditioning the generator on the deployment
context; in the \emph{random-prompt} configuration, prompts are drawn
from an unrelated distribution to obtain a control condition with the
same generator but without semantic grounding.

\paragraph{Structural and appearance conditioning}
For each rendered image $i\in\trainDataset$ we extract a monocular
depth map $d$ (using an off-the-shelf zero-shot estimator such as
MiDaS~\cite{Ranftl2022}), a Canny edge map $c$~\cite{canny1986edge},
and the ground-truth segmentation mask $m$ available from the renderer.
$d$ and $c$ are provided to ControlNet~\cite{zhang2023adding} to
constrain scene geometry, while $i$ is used by
IP-Adapter~\cite{ye2023ipadapter} to condition object appearance in
\acrshort{SDXL}~\cite{podell2023sdxl}. The caption $p$ conditions the
semantic content of the generated background.

\paragraph{Object composition}
The generator produces an augmented image $i'$. The target objects are
then re-inserted onto $i'$ using the segmentation mask $m$, producing
the final training image and preserving the original bounding-box and
class annotations. This mask-based copy-paste follows the practice
of~\cite{ghiasi2021copypaste}; a discussion of the resulting
photometric artifacts is deferred to
\cref{subsec:compositing}.

\begin{algorithm}[t]
\caption{S-GDR image generation.}
\label{alg:sgdr}
\begin{algorithmic}[1]
\STATE \textbf{Input:} Reference images $\refDataset$;
        rendered images $I$ with depth maps $D$, Canny edges $C$
        and segmentation masks $M$.
\STATE \textbf{Output:} Annotated augmented training images.
\STATE Caption $\refDataset$ with Qwen2-VL $\rightarrow$ prompt set $P$.
\STATE Initialize SDXL with ControlNet and IP-Adapter.
\FOR{$(i,d,c,m)\in(I,D,C,M)$}
    \STATE Sample $p\in P$ (or from a random distribution for the
           control condition).
    \STATE IP-Adapter encodes $i$ as a visual prompt for appearance.
    \STATE ControlNet encodes $(d,c)$ as spatial conditions.
    \STATE Generate augmented image $i' \leftarrow \text{SDXL}(p,d,c,i)$.
    \STATE Paste target objects from $i$ onto $i'$ using $m$.
    \STATE Save $i'$ with the original annotations of $i$ to
           $\trainDataset$.
\ENDFOR
\end{algorithmic}
\end{algorithm}

\subsection{Compared Configurations}
\label{sub:baselines}
All configurations share the same YOLOv8 detector, the same
$\trainDataset$ pool, the same $\refDataset$ context source, and the
same total training-set size of 200 images:

\begin{itemize}
    \item \textbf{Render baseline}: 200 images drawn at random from
    $\trainDataset$ (pure \acrshort{DR}).
    \item \textbf{Brightness filtering}: 200 images selected from
    $\trainDataset$ based on average-luminance similarity to
    $\refDataset$~\cite{martinez2024scap}.
    \item \textbf{Perceptual hashing}: 200 images selected from
    $\trainDataset$ based on low-frequency perceptual-hash similarity
    to $\refDataset$~\cite{martinez2024scap}.
    \item \textbf{GDR CycleGAN style transfer}: 100 random images from
    $\trainDataset$, augmented by unpaired CycleGAN~\cite{zhu2020unpairedimagetoimagetranslationusing}
    style transfer from $\refDataset$, combined with 100 random
    $\trainDataset$ images.
    \item \textbf{GDR random prompts}: 100 random images augmented via
    the pipeline of \cref{alg:sgdr} but with random (non-grounded)
    prompts, combined with 100 random $\trainDataset$ images.
    \item \textbf{\acrshort{S-GDR} guided prompts (ours)}: 100 random
    images augmented via the pipeline of \cref{alg:sgdr} with
    Qwen2-VL-generated prompts grounded on $\refDataset$, combined
    with 100 random $\trainDataset$ images.
\end{itemize}
Because the two right-most configurations share every component except
the source of the text prompt, their comparison isolates the effect of
\emph{semantic grounding} but does \emph{not} constitute a
leave-one-component-out ablation of the ControlNet or the IP-Adapter;
we make this scope explicit in \cref{sec:discussion,sec:limitations}.

% ==========================================================================
\section{Experiments and Results}
\label{sec:results}
% ==========================================================================

\begin{figure}[t]
\centering
\includegraphics[width=1\linewidth]{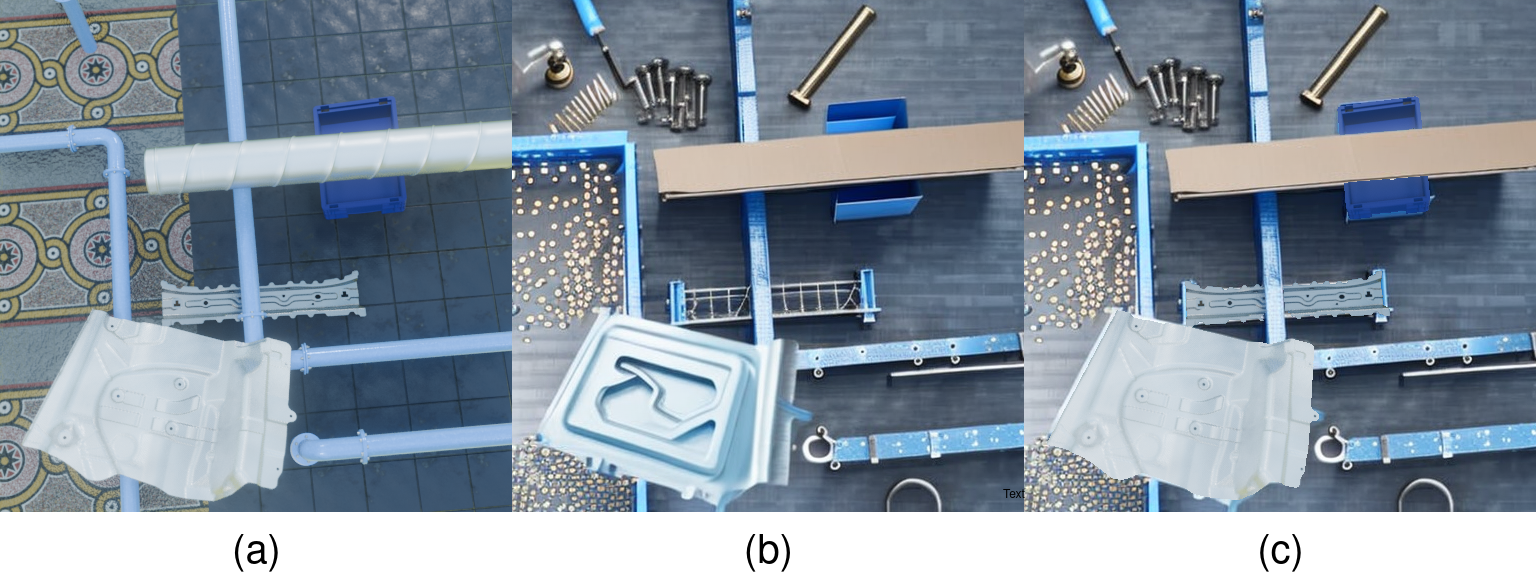}
\caption{Sample images produced by the \acrshort{S-GDR} pipeline.
(a) Original domain-randomized synthetic image.
(b) Generatively augmented background.
(c) Final training image with semantically contextualized background
and original target objects re-inserted via mask-based composition.}
\label{fig:guided_sd}
\end{figure}

\cref{fig:guided_sd} illustrates the output of the \acrshort{S-GDR}
pipeline: \cref{fig:guided_sd}(a) shows a rendered synthetic input from
$\trainDataset$; \cref{fig:guided_sd}(b) shows the \acrshort{SDXL}
output conditioned via ControlNet, IP-Adapter, and the contextual
prompt generated by Qwen2-VL from $\refDataset$;
\cref{fig:guided_sd}(c) shows the final training image after
mask-based composition of the target objects from (a) onto (b).

\begin{figure}[t]
    \centering
    \includegraphics[width=1\columnwidth]{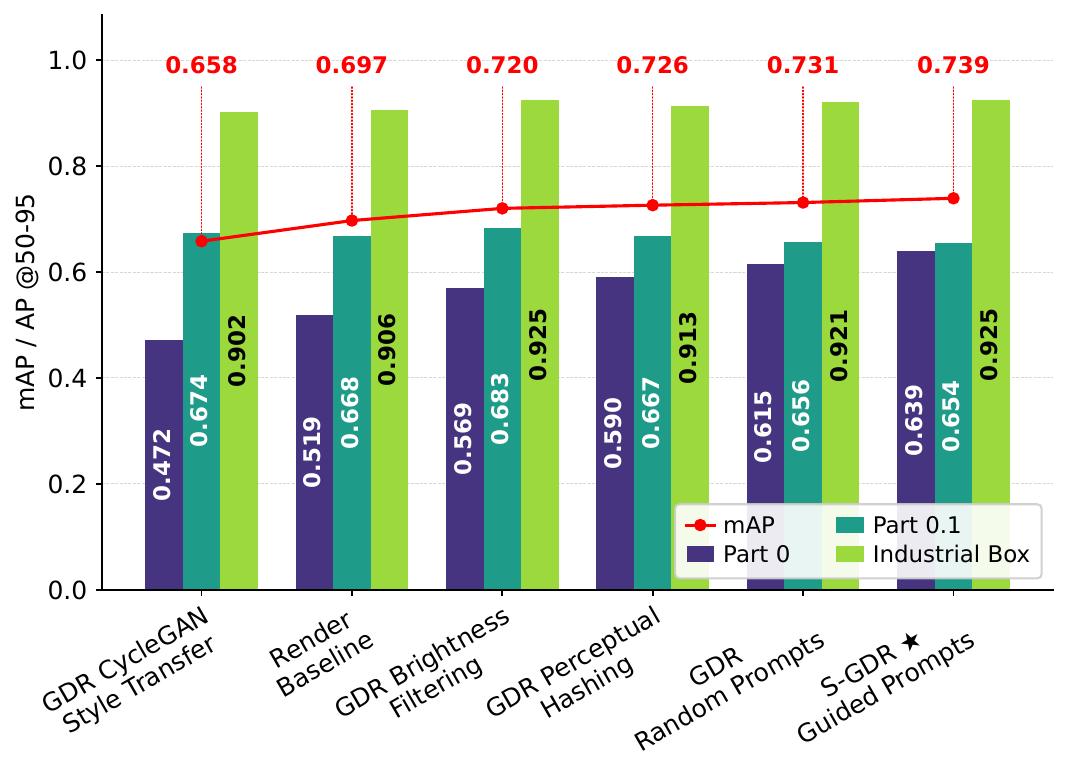}
    \caption{Per-class \acrshort{AP} and overall \acrshort{mAP}
    ($\mAPfull$) on the real test set $\testDataset$ of the automotive
    benchmark of~\cite{martinez2024scap}, for the six configurations
    of \cref{sub:baselines}. YOLOv8 was trained on 200 synthetic
    images per configuration. Bars: per-class \acrshort{AP}; red
    line: overall $\mAPfull$. Scores are single-run values.}
    \label{fig:200_img_map_results}
\end{figure}

\cref{fig:200_img_map_results} reports per-class \acrshort{AP} and
overall $\mAPfull$ on the real test set $\testDataset$, all with
training set size $|\trainDataset|=200$.

\subsection{Overall Results}
The render baseline establishes the floor with
$\mAPfull=0.697$. Feature-based selection improves over this baseline
to $0.720$ (brightness filtering) and $0.726$ (perceptual hashing) by
biasing the training subset toward the low-level statistics of
$\refDataset$. Contextualizing $\trainDataset$ with unpaired CycleGAN
style transfer from $\refDataset$ falls \emph{below} the baseline
($0.658$), which is consistent with the interpretation that unpaired
style transfer introduces features that do not correlate with the
target distribution.
\acrshort{S-GDR} with random prompts (unguided diffusion background
augmentation) improves the baseline to $0.731$, and \acrshort{S-GDR}
with guided prompts anchored on $\refDataset$ achieves the highest
score, $\mAPfull=0.739$.

Taken at face value, these observations are consistent with the
hypothesis that high-level semantic grounding of the augmentation
via a \acrshort{VLM} provides a stronger guidance signal than
low-level feature matching or unguided generation in the 200-image
regime evaluated here. Because the scores are single-run values, we
report a $+0.042$~$\mAPfull{}$ improvement over the render baseline
and a $+0.008$~$\mAPfull{}$  improvement over the strongest non-semantic
generative variant (random prompts) as observed magnitudes rather
than as statistically significant differences. Statistical
significance and cross-seed variability are discussed as open items
in \cref{sec:limitations}.

\subsection{Per-Class Observations}
\label{sub:per_class}
\cref{fig:200_img_map_results} shows that the improvement over the
render baseline is not uniform across classes. For the visually
most challenging class (\emph{Part~0}), \acrshort{S-GDR} with guided
prompts improves \acrshort{AP} substantially over the render baseline
(from $0.519$ to $0.639$). For the \emph{Industrial Box} class,
which already reaches $0.906$ AP with the render baseline, all
methods saturate in a narrow band ($0.902$--$0.925$), so absolute
improvements are limited.
For the intermediate class \emph{Part~0.1}, \acrshort{S-GDR} with
guided prompts and CycleGAN show slightly lower \acrshort{AP} than
the render baseline. A likely explanation is that the
context-injection process modifies mid-frequency background statistics
in a way that YOLOv8 has to ``compete'' with when localizing
mid-difficulty parts; a systematic verification of this hypothesis
requires per-class ablation studies that are beyond the scope of this
work-in-progress paper.

% ==========================================================================
\section{Discussion}
\label{sec:discussion}
% ==========================================================================

\subsection{What the Prompt Comparison Can and Cannot Attribute}
\label{subsec:ablation_discussion}
The gap between the \emph{random-prompt} and \emph{guided-prompt}
configurations isolates the effect of semantic grounding
\emph{conditional on} the presence of ControlNet, IP-Adapter, SDXL,
and mask-based composition. It does not attribute performance to
those components individually. A full leave-one-component-out
ablation, in which each conditioning branch (VLM prompt, ControlNet,
IP-Adapter) is disabled in turn, is the natural next experiment and
is called out explicitly in \cref{sec:limitations,sec:conclusion}.
Similarly, the effect of alternative captioners (e.g.,
LLaVA~\cite{liu2023llava}) and of prompt-engineering choices is not
resolved by the current data.

\subsection{Compositing Artifacts and Blending Alternatives}
\label{subsec:compositing}
The mask-based composition of \cref{alg:sgdr} inherits the well-known
limitations of copy-paste augmentation~\cite{ghiasi2021copypaste}:
photometric discontinuities at the object boundary
(lighting direction mismatch, shadow inconsistencies, color-tone
gaps) and, potentially, scale or perspective inconsistencies between
foreground and generated background.
Two families of blending techniques are commonly used to attenuate
these artifacts:
(i)~gradient-domain seamless
cloning~\cite{perez2003poisson}, which imposes source-gradient
constraints while enforcing continuity with the background, and
(ii)~learned image
harmonization~\cite{tsai2017deepharmonization}, which adapts the
foreground statistics to the background using a data-driven model.
Neither has been evaluated in the present pipeline, and a systematic
comparison is left for future work. Qualitative inspection of the
generated composites in \cref{fig:guided_sd} does not reveal
gross artifacts at typical viewing scales, but a formal artifact
assessment would require a labeled corpus of composited images that
we have not collected.

\subsection{Computational Cost, Scalability, and Deployment}
\label{subsec:cost}
A quantitative cost characterisation is out of scope for this WiP
paper, but we can describe the qualitative structure of the cost.
The pipeline of \cref{alg:sgdr} has three cost drivers:
(i)~Qwen2-VL captioning of $\refDataset$, which is executed once and
whose cost scales linearly with $|\refDataset|$;
(ii)~depth and edge extraction, which is deterministic and negligible
compared to (iii);
(iii)~SDXL generation with ControlNet and IP-Adapter, which
dominates and whose cost scales linearly with the number of
augmented images, i.e. with $|\trainDataset|/2$ in our configuration.
Compared to a pure render baseline, S-GDR adds one diffusion pass per
augmented image, which is orders of magnitude more expensive than a
single rendering call. However, the target regime is exactly the
one in which the training set is bounded by design (200 images in
our experiments), so the absolute cost of augmentation is bounded a
priori.
For \acrshort{HMLV} deployment, the amortized cost per retraining
cycle is dominated by the diffusion passes and can, if needed, be
mitigated by using the VLM-derived scene descriptions to condition
the randomization parameters of rendering engines such as
SynthRender~\cite{arayamartinez2026synthrenderirisopensourceframework}
directly, avoiding the diffusion step at the cost of losing some of
the appearance realism. A formal wall-clock, GPU-memory, and energy
characterisation is called out in \cref{sec:conclusion}.

% ==========================================================================
\section{Limitations and Potentials}
\label{sec:limitations}
% ==========================================================================
The evidence presented in this paper is subject to the following
scope conditions, each of which we make explicit:
\begin{enumerate}
    \item \emph{Single benchmark}: results are reported on the
    automotive benchmark of~\cite{martinez2024scap} only.
    Generalisation to other industrial domains is a research
    hypothesis, not an established fact.
    \item \emph{Single detector}: only YOLOv8 is used, with default
    hyperparameters. Whether the observed ranking transfers to other
    architectures (e.g., transformer-based detectors) is not resolved
    by this evaluation.
    \item \emph{Single training run per configuration}: all reported
    scores in \cref{fig:200_img_map_results} correspond to a single
    training run. Because SDXL generation and VLM captioning are
    stochastic, the $+0.042$ and $+0.008$ $\mAPfull{}$ gaps
    highlighted above should be treated as observed magnitudes;
    their statistical significance would require multiple training
    runs with different generation seeds and reporting mean$\pm$std.
    \item \emph{No component-wise ablation}: the guided- vs.
    random-prompt comparison does not attribute the effect to
    individual components. A leave-one-component-out ablation
    (VLM prompt, ControlNet, IP-Adapter) is needed to isolate their
    contributions.
    \item \emph{Dependency on a small real reference set}: even
    though training is annotation-free, the pipeline still needs
    tens of unlabeled real deployment images to build $\refDataset$
    and a small labelled $\testDataset$ for evaluation.
    \item \emph{VLM non-determinism}: hallucinations and off-domain
    lexical choices of Qwen2-VL~\cite{wang2024qwen2vl} can propagate
    semantic inconsistencies through the augmentation pipeline;
    quantifying this failure mode is left for future work.
    \item \emph{Compositing artifacts}: mask-based copy-paste can
    introduce photometric and geometric inconsistencies at object
    boundaries; blending alternatives such as Poisson
    editing~\cite{perez2003poisson} and learned
    harmonization~\cite{tsai2017deepharmonization} have not been
    evaluated.
    \item \emph{Scope to data-scarcity}: as
    \cite{araya-martinez2025genai,martinez2024scap} show, with
    sufficient rendering variability (an order of magnitude more
    training images) simpler feature-based methods can reach higher
    $\mAPfull{}$ at lower compute cost. \acrshort{S-GDR} is therefore
    positioned specifically for the 200-image regime, not as a
    general replacement for feature-based
    \acrshort{S2R} adaptation.
\end{enumerate}

Despite these open items, \acrshort{S-GDR} offers a principled path
toward fully automated synthetic-data contextualisation driven solely
by unannotated real reference images, removing the iterative human
tuning required by conventional \acrshort{DA} approaches.
\acrshort{VLM}-generated semantic descriptions operate at a level of
abstraction beyond the low-level color, texture, and structural
features employed by existing evaluators, and can also be reused
outside diffusion-based synthesis to condition classical rendering
engines directly, extending the utility of \acrshort{S-GDR} beyond the
data-scarcity regime.

% ==========================================================================
\section{Conclusions and Future Work}
\label{sec:conclusion}
% ==========================================================================
This work-in-progress paper evaluated \acrshort{S-GDR}, a semantic
\acrshort{DR} pipeline combining Qwen2-VL captioning of a small real
reference set with SDXL$+$ControlNet$+$IP-Adapter background synthesis
and mask-based object composition, on the automotive benchmark
of~\cite{martinez2024scap}.
With a fixed budget of 200 synthetic training images and YOLOv8 at
default hyperparameters, S-GDR with guided prompts reached
$\mAPfull=0.739$ on the real held-out test set, above a
domain-randomized render baseline of $0.697$, feature-based selection
variants ($0.720$ and $0.726$), CycleGAN style transfer ($0.658$), and
the same generative pipeline with random prompts ($0.731$).
These are observed magnitudes rather than statistically established
gaps.

As already established in~\cite{araya-martinez2025genai}, simpler
feature-based methods remain preferable when sufficient rendering
variability is available; the practical value of \acrshort{S-GDR} is
therefore specifically in the extreme data-scarcity regime, where
efficient training on small datasets is as important as terminal
accuracy.

Future work will address the open items enumerated in
\cref{sec:limitations}. In particular, we plan
(i)~a leave-one-component-out ablation of the VLM prompt, ControlNet
and IP-Adapter branches;
(ii)~a multi-seed evaluation with reported mean$\pm$std over at
least three independent training runs;
(iii)~an evaluation on an additional industrial domain to test
external validity;
(iv)~a controlled comparison of mask-based copy-paste, Poisson
blending~\cite{perez2003poisson} and learned
harmonization~\cite{tsai2017deepharmonization};
(v)~a quantitative wall-clock, GPU-memory, and energy
characterisation of the pipeline; and
(vi)~the unification of \acrshort{S-GDR} with
\acrshort{SAL}~\cite{ZHU202668}, coupling VLM-driven semantic
feedback with model-guided data generation to serve higher-value
synthetic training data across the full spectrum of data
availability.

% ==========================================================================
\bibliographystyle{IEEEtran}
\bibliography{IEEEexample}

\end{document}